# An Efficient Confidence Measure-Based Evaluation Metric for Breast Cancer Screening Using Bayesian Neural Networks

Anika Tabassum
Data Analytics
Ryerson University, Toronto, ON, Canada
anika.tabassum@ryerson.ca

Naimul Khan
Electrical, Computer and Biomedical Engineering
Ryerson University, Toronto, ON, Canada
n77khan@ryerson.ca

*Abstract—* Screening mammograms is the gold standard for detecting breast cancer early. While a good amount of work has been performed on mammography image classification, especially with deep neural networks, there has not been much exploration into the confidence or uncertainty measurement of the classification. In this paper, we propose a confidence measure-based evaluation metric for breast cancer screening. We propose a modular network architecture, where a traditional neural network is used as a feature extractor with transfer learning, followed by a simple Bayesian neural network. Utilizing a two-stage approach helps reducing the computational complexity, making the proposed framework attractive for wider deployment. We show that by providing the medical practitioners with a tool to tune two hyperparameters of the Bayesian neural network, namely, fraction of sampled number of networks and minimum probability, the framework can be adapted as needed by the domain expert. Finally, we argue that instead of just a single number such as accuracy, a tuple (accuracy, coverage, sampled number of networks, and minimum probability) can be utilized as an evaluation metric of our framework. We provide experimental results on the CBIS-DDSM dataset, where we show the trends in accuracy-coverage tradeoff while tuning the two hyperparameters. We also show that our confidence tuning results in increased accuracy with a reduced set of images with high confidence when compared to the baseline transfer learning. To make the proposed framework readily deployable, we provide (anonymized) source code with reproducible results at https://git.io/JvRqE.

## I. INTRODUCTION

Breast cancer is the most common cancer among women around the world according to World Health Organization [1]. The key to breast cancer control is early detection to improve breast cancer outcome and survival [1]. Mammography is the most common screening technology for breast cancer. It is a type of imaging that uses a low-dose X-ray system to examine the breast and is the most reliable method for screening breast abnormalities [2] before they become clinically perceptible. Screening mammography is done for detecting breast cancer. However, one big challenge here is low contrast in mammogram images, which makes it hard for radiologists to interpret the results [3]. Therefore, the use of computer aided diagnosis (CAD) has been on the rise for breast cancer screening [2][4].

To accomplish this, we have seen the usage of traditional approaches based on heavy feature engineering, as well as recent approaches based on deep convolutional neural networks. However, for a crucial task like cancer image screening, just classifying an image to a particular class (e.g. benign or malignant) is not enough, because it lacks any confidence or uncertainty measure associated with classification [5]. For example, if an image is classified as malignant, the radiologist might be interested in knowing how confident the CAD system is that it is malignant.

The target of this work is not only to compute such uncertainty measure in an efficient manner, but also to provide the radiologists with a tool to effectively control the accuracy-coverage tradeoff (explained in detail in section IIIB). We first train a deterministic point-estimate neural network using the pretrained ResNet-18 architecture with some modifications, thus leveraging transfer learning. For saving computational resources, we separate the feature extractor from this deterministic network to generate lower dimensional features and feed those to a separate smaller network which acts as our Bayesian neural network. Having computed the posterior distributions by applying Stochastic Variational Inference (SVI) [7], we introduce two tunable parameters **N** (sampled network fraction) and **P** (minimum probability) which together (both are explained in detail in section IIIB) can be used as a confidence measure and can be tuned to adjust the confidence level. We also demonstrate that higher confidence results in lower coverage of the classification, i.e., some images are rejected due to lack of confidence. We propose that the tuple *(accuracy, coverage, N, P)* can be our new evaluation criterion where *(N, P)* is the confidence measure. We obtain the mammography images from CBIS-DDSM [3] where the classification task is essentially binary (benign vs malignant) and demonstrate our tool in effect. The overall approach can in general be applicable to any domain beyond medical imaging and any number of classes.

## II. RELATED WORK

*Tsochatzidis et al.* [8] performed a comparative study on applying CNNs for breast cancer diagnosis. They made use of Alexnet, VGG, GoogLeNet, Inception Networks and ResNet.They showed that under fine-tuning scenario, pretrained networks achieve superior performance over networks trained from scratch. *Agarwal et al.* [9] showed similar results with transfer learning on VGG16, ResNet50 and InceptionV3.

*Xi et al.* [10] performed binary classification of mammography images using transfer learning. They made use of AlexNet, VGGNet, GoogleLeNet and ResNet and showed that VGGNet achieves the best overall accuracy while ResNet performs best for computing class activation maps. Part of our proposed framework has adopted this strategy.

*Rampun et al.* [11] performed classification of mammographic microcalcification clusters with confidence levels. They studied distribution of classifiers' probability outputs and used it as an additional confidence level metric to indicate reliability. They concluded that in breast CAD systems, the accuracy or AUC metric alone does not provide a complete representation of reliability. Although they do not make use of Bayesian neural networks, our paper is greatly motivated by this work given that we are also looking for a confidence measure-based evaluation criterion for classification.

Although not particularly in the domain of medical imaging, *Harper and Southern* [12] showed a Bayesian deep learning framework for prediction emotion from heartbeats by introducing a tunable confidence measure. Their confidence measure is based on the percentage of the output distribution that lies within a given class zone.

In terms of adopting a confidence or uncertainty measure in the domain of medical image classification, the work of *Leibig et al.* [13] has some similarity to ours. They showed a method for capturing uncertainty in disease detection using drop-out based Bayesian neural networks. Their method of measuring Bayesian uncertainty was based on the recent finding [14] that a multi-layer perceptron with added dropout after every weight layer is mathematically equivalent to approximate variational inference [15] in the deep Gaussian Process model [16, 17], which holds for any number of layers and arbitrary non-linearities. They extended this idea to incorporate convolutional layers [18]. The uncertainty for a given test image was obtained by simply keeping the dropout mechanism switched on at test time and performing multiple predictions. They used the Messidor dataset [19] for their experiments and manifested a monotonic increase in prediction accuracy for decreasing levels of tolerated model uncertainty. They pointed out that one of their main motivations for resorting to a dropout-based Bayesian approach, as opposed to a Gaussian process (GP) approach was that while GPs theoretically seem more appealing, they scale badly with both the dimensionality of the feature space and the size of the dataset.

In our proposed framework, unlike [13] we rely on a Gaussian process (Stochastic Variational Inference) for our Bayesian posterior approximation, and then use a few hyper parameters that work on networks sampled from that posterior, to tune the level of uncertainty. For dealing with performance and scalability issues (the main criticism against GPs in [13]), we mainly rely on transfer learning, along with dividing our network architecture into a deterministic portion which acts as a lower-dimensional feature generator, and another relatively small neural network on which the actual Bayesian inference is performed. Dividing the network into a deterministic and a Bayesian portion while employing transfer learning makes our proposed method extremely modular, therefore giving it the ability to be plugged into existing CAD systems which are employed at hospitals or clinics. This modular approach is inspired by the work of *Riquelme et al.* [20], where they apply a Bayesian linear regression on the last layer of a deep neural network.

## III. PROPOSED FRAMEWORK

Our proposed framework has two main components – (i) A modular network architecture with a feature generator combined with a Bayesian network (ii) Tunable hyperparameters on the posterior distribution learnt by the Bayesian training to come up with a confidence measure.

### A. Modular Network Architecture

To achieve a proper Bayesian training, we first need to make sure we have a good architecture of a deterministic (non-Bayesian) neural network, because the Bayesian neural network will be based on the non-Bayesian one. There are already well studied and researched neural network architectures like AlexNet [21], ResNet [23], and VGG [22] (along with their pre-trained versions). There are two avenues to transform one of these architectures into a Bayesian network:

1. Try end-to-end Bayesian learning from scratch, not leveraging the pre-trained version of the chosen network architecture (i.e., not leveraging transfer learning). Initialize the weights and biases with random priors (e.g with normal distributions with zero mean and unit standard deviation) and then apply a Gaussian Process like Stochastic Variational Inference (SVI) end-to-end to learn the posterior distributions.

2. Try leveraging transfer learning and do Bayesian learning via the Gaussian process on top of that. We call this second approach a modular approach.

However, there are a few empirical problems associated with both of these approaches:

1. The problem with approach 1 above is time and resource complexity. For example, a network architecture like ResNet-18, with 3 additional fully connected layers, has over 11 million parameters, so learning posterior distributions via a Gaussian process for each of these parameters will be very time consuming could be impractical for wider deployment and re-training.

2. Approach 2, has not been investigated widely, other than [20] provides results on simple numerical datasets.

In order to get around these issues, we decided to modify approach 2 above to make it modular. The steps are outlined below:

1. Select a pretrained deep neural network (such as ResNet-18) and adapt it to leverage transfer learning. For example, for ResNet-18, replace the fully connected layer with a stack of trainable fully connected layers to leverage transfer learning.
2. Train this network deterministically (non-Bayesian) to reach a reasonable classification accuracy.
3. Now divide the network into 2 different independent networks – one will be a network containing a majority of the layers and convolutional blocks so that it can act as a lower dimensional feature generator, and the other will be a much simpler network consisting of a few fully connected layers, which can be used for end to end Bayesian learning using the lower dimensional features as input. For example, when using an adapted version of ResNet, the feature generator network will be the portion of the network before the fully connected layers start, and the smaller network for Bayesian learning will just be the fully connected layers followed by softmax ( e.g in our experiemnts it is a **512**-element feature vectors, as opposed to 224 x 224 images for ResNet). A similar breakdown has to be applied if any other network architecture is being used.

4. Now perform the Bayesian inference via a Gaussian Process such as SVI on the smaller network and with lower dimensional training data, to learn the posterior distribution of the parameters.

This modular approach makes the Bayesian learning process faster, since the Gaussian process (SVI) is now being applied to a much smaller network and the input data for this network is also lower dimensional. This Bayesian learning is easily achievable within a reasonable time frame using off-the-shelf tools (like Pyro [26]). Together, this end-to-end approach gives us a way to do Bayesian learning on top of leveraging transfer learning in an effective manner. This two-step approach is also attractive in the sense that popular neural network architectures are already being deployed at hospitals and clinics for computer-aided diagnosis. The Bayesian network can be an additional module to be tacked on to these existing architectures. **Fig. 1** summarizes the steps mentioned in this subsection.

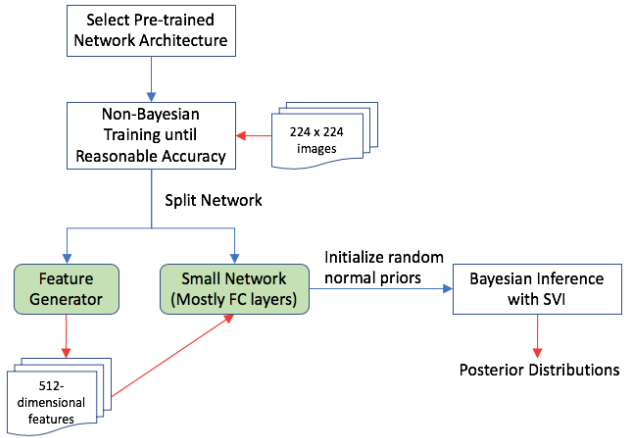

Fig. 1: Framework for Bayesian Posterior Inference

### B. Tunable Hyperparameters for Confidence Measurement

Once we have the posterior distributions of the network parameters (weights and biases) calculated from section IIIA, the next step is to come up with a confidence measure. Our proposed confidence measure consists of two tunable parameters. The steps to calculate these are described below:

1. Sample a reasonable number (1000 in our experiments) of networks from the posterior distributions using Monte Carlo sampling. Each of these sampled networks is a deterministic network by itself.
2. Classify each image by each of the sampled networks. Record the probabilities for both classes (benign and malignant) for each image.
3. Use two parameters **N** and **P**, where **N** denotes the fraction of the sampled number of networks that have a minimum probability **P** on a certain image being of a particular class (either benign or malignant).
4. With these two tunable parameters **N** and **P**, we can define a confidence measure. For example, if we have **1000** sampled networks, then **N = 0.6** and **P = 0.7** would mean at least **600** networks out of the **1000** have to have a probability of at least **0.7** for an image being either benign or malignant, otherwise the image will be skipped for classification. In other words, by incorporating both **N** and **P** in the confidence measure, we account for agreement among a portion of the sampled networks and find out how strongly each network feels about the classification.

Under the above settings, naïve expectation would be that as **N** and **P** go higher (higher confidence, lower uncertainty), we should be getting higher accuracy, while as N and P go lower (lower confidence, higher uncertainty), the accuracy should decrease. However, raising the value of **N** and **P** might also result in some images being skipped for classification. For example, consider a case where we have **1000** sampled networks, **N = 0.9** and **P = 0.9**, which demands that at least **900** out the **1000** networks must have a probability of at least **0.9** for an image being either benign or malignant. This might result in a number of images being skipped for classification, since we are demanding too high of a confidence. This is the case of *lower coverage*. At higher values of **N** and **P** (higher confidence), we will have lower coverage (many images skipped), but the accuracy on the covered images will be high. On the other hand, at lower values of **N** and **P** (lower confidence), we will have (*higher coverage* not too many images skipped) but the accuracy on the covered images would be lower. In short, tuning the values of **N** and **P** gives us a way to decide where in the *accuracy-coverage tradeoff* we want to settle. Therefore, the **N** and **P** values, along with the accuracy and coverage, comprise our new evaluation metric, which can be formalized as a tuple *(accuracy, coverage, N, P)* where **N** and **P** comprise the confidence measure.

Although in this paper we are using classification accuracy as the primary evaluation criterion, in general, the framework allows any evaluation criterion to be paired with a confidence measure. Other possible candidates include precision, recall, F1-score, ROC AUC score etc., some of which might make more sense for an imbalanced classification case. Ours being a relatively class-balanced classification case, we chose to go with accuracy for our task.

### IV. EXPERIMENTAL SETUP

#### A. Dataset

The data has been obtained from the CBIS-DDSM (Curated Breast Imaging Subset of DDSM) [3] dataset. For the purpose of our classification problem, we utilize the cropped Region-of-Interest (ROI) image patches containing the mass that are provided with the dataset. The ROI-cropped dataset contains 1696 labelled grayscale images. Of those, 912 are labelled as benign and 784 are labelled as malignant. Therefore, the dataset is reasonably class-balanced. **Fig. 2** shows samples of one benign and one malignant image patch.

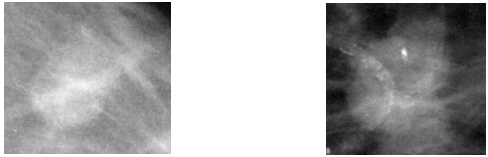

(a) Benign    (b) Malignant

Fig. 2: Sample benign and malignant images patches

### B. Data Preprocessing

*1) Image Resizing:* We have resized all images to 224 x 224 pixels to match with the input image size of popular deep networks.

*2) Normalization:* We have performed standard normalization on all the images with zero mean and unit standard deviation in order to reduce the chance of overfitting.

*3) Train-Validation Split:* Although CBIS-DDSM comes with a pre-division of the dataset into training/test, for more generalization, we merged these two datasets and then split them again, into our own training and validation datasets. This was done to ensure similar class ratio (benign to malignant) of the images in the training and validation sets. This splitting was done 80:20 training to validation ratio and in a stratified manner with regards to the class (benign vs malignant) distribution. We created 5 such stratified splits so that each such split can be used for cross-validation to test the generalization of our approach. For each of the 5 splits, we have 1357 training image and 339 validation images.

*4) Data Augmentation:* We have performed data augmentation by applying random rotation (from 0 to 360 degrees) and random vertical and horizontal flips on the images. Note that this was done for training data only. This augmentation was done on the fly during the training process, meaning that each epoch would have training images with different orientations, effectively increasing the training set size manifold (as opposed to the actual 1357 training images) and making the training process more robust.

### C. Network Architecture Selection and Adaptation

Referring to the framework described in section IIIA, the first step was to select and adapt a deep neural network architecture on which we would perform non-Bayesian training. We have experimented with pre-trained AlexNet [21], VGG-16 [22] and ResNet-18 [23]. To adapt these networks to a binary classification problem, we followed [10] and in each case (AlexNet, VGG-16 and ResNet-18), we have dropped the last fully connected layer and replaced that with our own block of trainable fully connected layers. We also made a few later convolutional blocks trainable while freezing the earlier part of the networks. This setting allows us to learn domain specific features on top of the pretrained networks. For the CBIS-DDSM dataset, ResNet-18 provided the best performance. For this reason, in this paper, we will confine the discussion to the results obtained by ResNet-18 only. We replaced the fully connected layer of ResNet-18 with 3 new fully connected layers. Also, of the four residual convolutional blocks, we made the latter 2, namely Conv3 and Conv4 as trainable. We froze the earlier part of the network. We also modified the softmax layer to have only 2 classes as per our need.

### D. Non-Bayesian Training

We trained via our **1357** training images (for each split separately and with random augmentation applied on them) we used a batch size of 8 and an initial learning rate of **0.0001** for the earlier part of the network (except for the fully connected part). For the fully connected part we gradually increased the initial learning rate layer by layer, **0.0001** for the first layer, **0.001** for the second one and **0.002** for the last one. We also used an exponential learning rate decay scheme and an Adam optimizer. The training was done using the PyTorch [25] framework on Google Colab platform which uses a Nvidia Tesla K80 GPU and took approximately 12 minutes to finish with **25** epochs. This provided a reasonable accuracy of **81**% on the validation set on average. Since achieving the best possible accuracy is not the focus of our work, we did not experiment with the parameters any further, and utilized this trained network with **81**% accuracy as a *baseline feature extractor* for our Bayesian network.

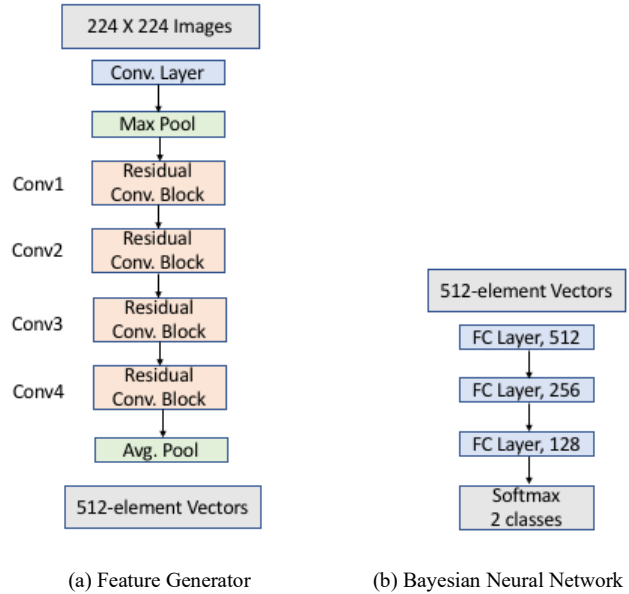

(a) Feature Generator    (b) Bayesian Neural Network

Fig. 3: Neural Network Architecture for Bayesian Inference

### E. Network Split

As per the framework from section IIIA, the network splitting is done so that we have a feature generator consisting of mainly convolutional blocks which takes in 224 x 224 sized images and generates 512-element feature vectors, and a separate small fully connected network which has just 3 fully connected layers followed by a softmax. Notice that this small network only has just over 164 thousand parameters as opposed to the over-11 million parameters in the full network, making the subsequent steps easily executable on commodity hardware. **Fig.** 3 summarizes the split.

### F. Bayesian Training

We re-initialized the weights of the small fully connected network with normally (Gaussian) distributed priors having zero mean and unit standard deviation. We then generated the

512- element features using the feature generator. Here we applied random augmentation in a way that we have 10 augmented images for every original training image (no augmentation was done for validation images). Since we have 1357 training images, we effectively ended up generating 13570 training images and correspondingly 13570 number of 512-element vectors from the feature generator. These 512-element vectors were fed to the small network and SVI was applied using Pyro software package [26] to learn the posterior distribution of the weights. The hyper-parameters (optimizer, learning rate) were the same as those we used during the non-Bayesian training, and we let it run for 10 epochs on the same Google Colab environment. The process took approximately 30 minutes to finish. The code to reproduce the results is available at the following anonymized Github repository:

https://github.com/ICHISubmission/Metric-Breast-Cancer

## V. EVALUATION AND RESULTS

As per the framework presented in section IIIB, we sampled 1000 networks from the posterior distributions and performed evaluation by tuning the values of N and P.

As a first check, we performed forced prediction by ignoring **N** and **P** and forcing Bayesian network to predict a class for every image by taking the average predicted probability for both classes for each image, and taking the class that has the higher average as prediction. This gives us **100%** coverage and a prediction accuracy of **81%**. Note that this is the same accuracy we got by using the single deterministic network in section IV, proving that our proposed network can also be converted to a traditional classifier if need be.

Next, we demonstrate the value of our control parameters **N** and **P**. We have **339** validation images on average. We found that when we set N above 0.9, all test images were skipped and the Bayesian network had no coverage, regardless of the value of **P**, which is expected, because it's unlikely that 95% of the networks would have a minimum probability for a class for any image. In general, if we kept **N** to a moderate value like 0.5 and then varied **P**, we noticed an upward trend in accuracy and a downward trend in coverage, which is expected. An almost similar trend shows up if we held **P** at 0.5 and varied **N** instead, which is also expected. The results have been summarized in **Table I** and **Fig. 4**. **Table Ia** and **Fig. 4a** capture the accuracy and coverage values and trends for varying **N**, keeping **P** at 0.5, while **Table Ib** and **Fig. 4b** capture the accuracy and coverage values and trends for varying **P**, keeping **N** constant at 0.5 (standard deviation is over the 5 splits of cross-validation).

Based on the observations, the conceptual expectation we laid out in section IIIB is verified. Thus, instead of only accuracy, a tuple *(accuracy, coverage, N, P)* can be our new evaluation criterion with *(N, P)* being the measure of confidence. We should point out that practically it might make more sense to keep **P** at a constant value (e.g. 0.5) and tune **N**, as opposed to keeping **N** constant and tuning **P**; this is because tuning **N** and keeping **P** constant effectively means varying the degree of polling on the sampled set of networks while expecting a fixed minimum classification probability agreement among the polled ones, whereas tuning **P** keeping **N** constant means freezing the polling space and varying the value of the minimum classification probability agreement in that frozen network space. In other words, it is **N** that allows us to choose how much of the sampled network distribution we will cover, and hence is a better measure of practical uncertainty.

We also compare our results with a baseline, where instead of picking the most confident test samples, we picked an equal number of test samples randomly. For example, the case where **P = 0.5** and **N = 0.85** in **Table Ia**, out of **339** images, only **28** images were classified with **96%** accuracy and the rest were skipped. In our baseline, instead of the Bayesian network, we take **10** different random samples of **28** images from the **339** validation images (from each split) and calculate the average accuracy of those samples by the non-Bayesian neural network (from section IVD).

TABLE I. ACCURACY AND COVERAGE VS N AND P VALUES

| N | P | Total Images | Skipped Images | Accuracy ±st. dev. | Coverage | Baseline Accuracy |
|---|---|---|---|---|---|---|
| 0.95 | 0.5 | 339 | 339 | NA | 0.00 | NA |
| 0.85 | 0.5 | 339 | 311 | 96±0.8 | 0.08 | 80 |
| 0.75 | 0.5 | 339 | 257 | 94±1 | 0.24 | 81 |
| 0.55 | 0.5 | 339 | 31 | 85±0.8 | 0.91 | 80 |
| 0.35 | 0.5 | 339 | 0 | 82±0.9 | 1.00 | 81 |
| 0.2 | 0.5 | 339 | 0 | 82±0.8 | 1.00 | 81 |

(a) Accuracy and Coverage vs N (at P = 0.5)

| N | P | Total Images | Skipped Images | Accuracy ±st. dev. | Coverage | Baseline Accuracy |
|---|---|---|---|---|---|---|
| 0.5 | 0.95 | 339 | 85 | 88±0.8 | 0.75 | 80 |
| 0.5 | 0.85 | 339 | 55 | 86±0.9 | 0.84 | 81 |
| 0.5 | 0.75 | 339 | 35 | 85±0.8 | 0.90 | 81 |
| 0.5 | 0.55 | 339 | 10 | 82±0.8 | 0.97 | 81 |
| 0.5 | 0.35 | 339 | 0 | 81±1 | 1.00 | 81 |
| 0.5 | 0.2 | 339 | 0 | 81±0.9 | 1.00 | 81 |

(b) Accuracy and Coverage vs P (at N = 0.5)

As can be seen in **Table I,** the accuracy of our proposed framework is always higher than the baseline. The baseline accuracy stays close to around **81%**, whereas the Bayesian accuracy clearly grows with lower coverage. The baseline trends are also shown in **Fig. 4a** and **4b** as dotted lines.

This proves that the increase in accuracy for the confident images is not just due to a small size effect. This is a clear demonstration of the confidence measure working as expected.

We should also point out that setting **N** and **P** to 0.5 while varying the other was a conscious choice to identify trends. A value too low (such as 0.2) is impractical and a value too high might be too aggressive. We have also created separate graphs for each pair of **N** and **P** values for all **(N, P)** pairs we used. A clear trend emerges only when one of them is set closer to 0.5. Due to lack of space we only present the best trends, which were found at 0.5.

The takeaway from the findings is that the parameters **N** and **P** can be tuned to a desirable level and the higher the values of these parameters, the higher confidence we will have in the predictions/classifications, in exchange for possibly lower coverage. For a domain like medical mammography image classification, it will be up to the mammography experts to determine what value of **N** and **P** is reasonable. After setting reasonable values for **N** and **P**, previously unseen images that will be classified by the Bayesian network (either as benign or malignant) are more likely to be correctly classified, whereas images that will be skipped and denied classification would need further investigation. Notice that the overall approach described in sections III and IV can in general be applied to any domain (beyond mammography) and any number of classes, and thus can be used as a skeleton framework for an uncertainty/confidence measurement task.

Due to resource constraints, we could not run a full Bayesian network identical to the modular architecture we have. It is highly likely that a full Bayesian network will result in increased accuracy, but with significantly higher computational cost. Hence, we put forward the idea of a modular architecture.

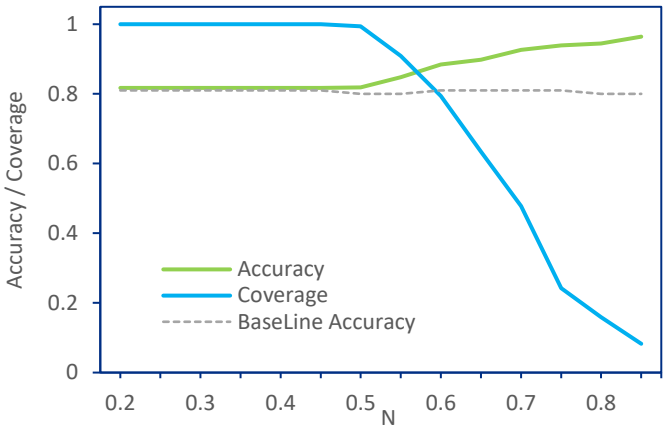

(a) Accuracy and Coverage vs N (at P = 0.5)

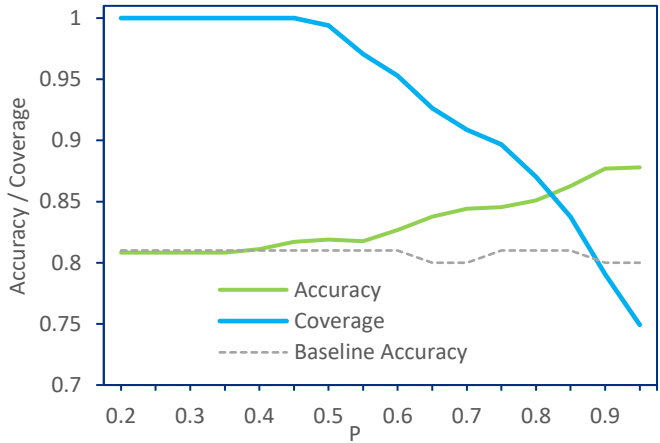

(b) Accuracy and Coverage vs P (at N = 0.5)

Fig. 4: Accuary and Coverage Trends vs N and P values

## VI. DISCUSSION AND FUTURE WORK

In this paper, we propose a confidence measure-based evaluation metric for computer-aided diagnosis systems. We propose a modular network architecture, where a traditional neural network (ResNet-18) is used as a feature extractor with transfer learning, followed by a simple Bayesian neural network. By utilizing a two-stage approach, we reduce the computational complexity significantly, which makes the proposed framework an attractive option for wider deployment, or simply as an add-on to existing systems that utilize deterministic networks. We show that by providing a tool to tune two hyperparameters of the Bayesian neural network, namely, fraction of sampled number of networks and minimum probability, the framework can be adapted to the liking of the domain experts. Finally, we argue that instead of just a single number such as accuracy, a tuple of accuracy, coverage, sampled number of networks, and minimum probability can be utilized as an evaluation metric of our framework. Experimental results provide an in-depth analysis of the effect of tuning the parameters on the accuracy-coverage tradeoff. We also compare with a baseline non-Bayesian network to show that our confidence tuning process can effectively filter out images with less confidence to increase accuracy.

During the training of our deterministic and Bayesian neural networks, we noticed that the more accurate the deterministic network is, the more consistent the behaviour of the Bayesian network is. This means that without a reasonably good feature extractor that we used as a precursor, the expected accuracy-coverage trade-off behaviour cannot be observed. It was only after we achieved over 80% accuracy through the deterministic neural network (and hence the feature generator) that the expected upward and downward trends in Fig. 4 were achieved via the Bayesian neural network. This implies that before a Bayesian approach can be relied upon, we first need to train a good network architecture with proper hyper parameter tuning. We settled on ResNet-18. However, more complex network architectures can be explored including some in the ResNet family (ResNet-50, ResNet-152). During transfer learning, we also noticed that the more layers we make trainable, the accuracy generally increases. With a more complex network architecture, tuning how many layers to freeze would be one interesting avenue to explore.

The prior distributions we assigned to the Bayesian layer parameters were simple Gaussian distributions with zero mean and unit standard deviation. Ideally, these priors should come from a more rigorous estimation, possibly based on the nature of the input data along with domain knowledge. Work such as [24] can be investigated for this purpose.

To make the approach computationally achievable with limited available resources, we performed training in two separate stages for the feature extractor and the Bayesian network. An end-to-end Bayesian learning approach on top of transfer learning to keep the computational requirement manageable is something we would be exploring in future. In that case, the priors to be assigned to the Bayesian layer parameters could come from the parameter values learnt for deterministic training.